\documentclass[10pt,journal,compsoc]{IEEEtran}
\usepackage{graphicx}
\usepackage{amsmath,amssymb} 
\usepackage{epsfig}
\usepackage{amsmath, bm}
\usepackage{amssymb}
\usepackage[normalem]{ulem}
\useunder{\uline}{\ul}{}
\usepackage{nicefrac}
\usepackage{comment}
\usepackage{enumitem}
\usepackage{adjustbox}
\usepackage{mathabx}
\usepackage{tabu}
\usepackage{booktabs}
\usepackage{diagbox}
\usepackage{cuted}
\usepackage{capt-of} 
\usepackage[table,x11names]{xcolor}
\usepackage{microtype}
\usepackage{multirow}
\usepackage{csquotes}
\usepackage{url}

\newcommand{\etal}{\textit{et al}. }

\ifCLASSOPTIONcompsoc
  \usepackage[nocompress]{cite}
\else
  \usepackage{cite}
\fi

\begin{document}

\title{Fully Dynamic Inference with Deep Neural Networks} 

\author{Wenhan Xia, Hongxu Yin, Xiaoliang Dai,
        and~Niraj K. Jha,~\IEEEmembership{Fellow,~IEEE}
        
\IEEEcompsocitemizethanks{\IEEEcompsocthanksitem
This work was supported by NSF under Grant No. CNS-1907381.
W. Xia and N. K. Jha are with the Department of Electrical Engineering, Princeton University, 
Princeton, NJ, 08540.
\IEEEcompsocthanksitem H. Yin is with NVIDIA.%
\IEEEcompsocthanksitem X. Dai is with Facebook.
}}

\IEEEtitleabstractindextext{%
\begin{abstract}
Modern deep neural networks are powerful and widely applicable models that extract task-relevant 
information through multi-level abstraction. Their cross-domain success, however, is often achieved 
at the expense of computational cost, high memory bandwidth, and long inference latency, which 
prevents their deployment in resource-constrained and time-sensitive scenarios, such as edge-side 
inference and self-driving cars. While recently developed methods for creating efficient deep neural 
networks are making their real-world deployment more feasible by reducing model size, they do not 
fully exploit input properties on a per-instance basis to maximize computational efficiency and 
task accuracy. In particular, most existing methods typically use a one-size-fits-all approach that 
identically processes all inputs. Motivated by the fact that different images require different 
feature embeddings to be accurately classified, we propose a fully dynamic paradigm that imparts 
deep convolutional neural networks with hierarchical inference dynamics at the level of layers and 
individual convolutional filters/channels. Two compact networks, called Layer-Net (L-Net) and 
Channel-Net (C-Net), predict on a per-instance basis which layers or filters/channels are redundant 
and therefore should be skipped. L-Net and C-Net also learn how to scale retained computation outputs 
to maximize task accuracy. By integrating L-Net and C-Net into a joint design framework, called 
LC-Net, we consistently outperform state-of-the-art dynamic frameworks with respect 
to both efficiency and classification accuracy. On the CIFAR-10 dataset, LC-Net results in up to 
$11.9\times$ fewer floating-point operations (FLOPs) and up to $3.3\%$ higher accuracy compared to 
other dynamic inference methods. On the ImageNet dataset, LC-Net achieves up to $1.4\times$ fewer FLOPs and up to $4.6\%$ higher Top-1 accuracy than the other methods.
\end{abstract}

}

\maketitle
\IEEEdisplaynontitleabstractindextext
\IEEEpeerreviewmaketitle

\IEEEraisesectionheading{\section{Introduction}\label{sec:introduction}}

\IEEEPARstart{I}{n} recent years, deep neural networks (DNNs) have dramatically accelerated the field of artificial intelligence. Their ability to represent data through increasingly more abstract layers of feature representations has proven effective in numerous application areas, such as image classification, speech recognition, disease diagnosis, and neural machine translation~\cite{AlexNet, deepspeech2, speechlstm, diabdeep, translation}.  With increased access to powerful computational resources and large amounts of labeled training data, e.g., ImageNet has 1.28 million images from 1,000 different categories~\cite{imagenet}, DNNs can achieve super-human performance on a variety of tasks.
\\\\
One strategy for increasing DNN expressivity and accuracy relies on adding more layers and 
convolutional filters. The evolution of DNNs over the past few years exemplifies this approach. For 
example, the ImageNet based DNN champion VGG-19~\cite{vgg} is $2.4\times$ deeper and has
$27.5\times$ more floating-point operations (FLOPs) than the previous champion AlexNet, and
DeepSpeech2 \cite{deepspeech2} is 
$2\times$ deeper and requires $5.8\times$ more computation than the preceding DeepSpeech. Despite 
their improved accuracy, these bulkier models are unsuitable for edge-side inference, which typically 
faces stringent latency and memory constraints. 
\\\\
To address the deployment limitations for modern DNNs, an emerging stream of research focuses on 
increasing model compactness. One set of approaches aims to learn compact architectures. For example, 
neural architecture search (NAS) uses gradient information to unveil new building 
blocks~\cite{rl3,nasnet}. Model 
compression methods, such as pruning, are used widely to reduce the computational overhead in 
DNNs~\cite{han2015deep}. The derived compact architectures offer computational savings with negligible
accuracy degradation. An alternative approach focuses on reducing computational overhead through 
weight quantization, where bit precision can be reduced without affecting accuracy~\cite{zhuternary,zeroq}.
\\\\
One property common to these methods is that the same model processes different input instances. 
Given the fact that different instances have unique visual features, a natural question arises: does every instance require all levels of embeddings and the same set of feature maps to be accurately 
classified? Intuitively, deeper embeddings may not be necessary for easily classifiable images. 
Therefore, to maximize computational efficiency, the additional computations associated with deeper 
layers should be reserved only for difficult input instances. In addition, since convolutional 
channels/filters capture class-specific features, unnecessary computations can be saved by skipping
irrelevant channels during inference. \\\\
Dynamic inference is an emerging approach that exploits input properties to selectively execute 
salient subsets of computations needed for accurate classification. Unlike static methods that 
permanently remove neurons to improve model efficiency, dynamic approaches only transiently suppress 
computations depending on the input instance. Thus, dynamic methods maximize efficiency and preserve 
model expressivity. To date, however, these methods have resulted in models with limited dynamics. As 
a result, they cannot fully adapt to the computational needs of each instance. For example, filter 
pruning approaches operate at the filter level but not the layer level, which may result in 
the execution of redundant computations and hence incur unnecessary computational cost. 
In addition, these approaches often use reinforcement learning to make selection decisions, which
is computationally intensive.
\\\\
In this work, we propose a framework that offers fully dynamic inference on a per-instance basis. At 
the core of our approach lies the design of predictive control nets placed parallel to a backbone 
network. Specifically, we introduce two novel auxiliary networks, Layer-Net (L-Net) and
Channel-Net (C-Net), that, respectively, assist with dynamic layer and channel skipping and scaling.The joint topology consisting of both control nets, referred to as LC-Net, enables fully dynamic inference that simultaneously improves inference efficiency and accuracy by (1) determining which channels and blocks to execute at 
inference time, (2) scaling retained channels and blocks with a salience score to maximize accuracy, 
and (3) predicting salient computations on-the-fly without halting the inference flow or incurring latency overhead.
\\\\
We summarize our contributions as follows:
\begin{itemize}
    \item We propose an instance-based fully dynamic inference paradigm. The method enables, for the first time, fully flexible model depth and width at inference time. It is completely compatible with existing training pipelines and results in minimal overhead and high accuracy.
    \item We introduce, study, and verify an on-the-fly predictive skipping and scaling mechanism, which incurs no halting time during inference.
    \item We introduce the ReLU-1 activation for predictive salience generation that can (1) zero out layers and channels to reduce computational cost and (2) re-scale the remaining layers and channels to improve accuracy.
    \item We qualitatively and quantitatively validate our approach on CIFAR-10 and ImageNet datasets. Both sets of experiments show that our approach results in state-of-the-art performance in terms of model efficiency and accuracy.
    \end{itemize}

\section{Related Work}
In this section, we discuss methods for deriving efficient neural networks, which can be broadly categorized into static and dynamic approaches. We provide a detailed overview of both categories. 

\subsection {Static Approaches}
Most DNNs are computationally-intensive and over-parameterized~\cite{hanpruning}. Several approaches
have been proposed to create efficient DNNs, including the design of novel compact neural network architectures and compression of existing models.  We summarize these approaches next.
\\\\
\textbf{Compact architecture design}:
Exploiting efficient building blocks and operations can significantly reduce DNN computational cost. For example, MobileNetV2 shrinks the model size and reduces the number of FLOPs with inverted residual building blocks~\cite{mobilenetv2}. Ma \etal propose another compact convolutional neural network (CNN) architecture that uses channel shuffle operation and depth-wise 
convolution~\cite{shufflenetv2}.  Wu \etal suggest replacing the spatial convolution layers with 
shift-based modules that have zero FLOPs. The generated ShiftNet has substantially less computational 
and storage costs~\cite{shift}. In addition, automated compact architecture design provides a 
promising solution~\cite{mnasnet}. Dai \etal propose an automated architecture adaptation and search framework based on efficient performance predictors \cite{chamnet}.
\\\\
\textbf{Model compression}:
Apart from compact architecture design, compressing and simplifying existing models has emerged as a promising approach~\cite{scann}. Network pruning is a successful DNN compression technique that removes redundant connections and neurons. Han \etal have shown that the number of parameters of
VGG-16 can be reduced by more than $13\times$ without any accuracy loss~\cite{hanpruning}. A recent 
work combines pruning with network growth and improves the compression ratio of VGG-16 by another 
2.5$\times$~\cite{nest}. Furthermore, structured sparsity and pruning can significantly reduce run-time latency~\cite{structuredsparsity}. Low-bit quantization is another powerful tool for reducing storage cost~\cite{han2015deep}. Zhu \etal show that replacing a full-precision (32-bit) 
weight representation with ternary weight quantization only incurs a minor accuracy loss for 
ResNet-18 but significantly reduces the storage and memory costs~\cite{zhuternary}.  
\\\\
\textbf{Knowledge distillation:} Knowledge distillation allows a compact student network to distill information (or $``$dark knowledge$"$) from a more accurate, but 
computationally-intensive, teacher network (or a group of teacher networks) by mimicking the 
prediction distribution, given the same data inputs. The idea was first introduced by 
Hinton \etal \cite{hinton2015distilling}. Since then, knowledge distillation has been used to discover efficient networks. Romero \etal propose FitNets that distill knowledge from the teacher's hint layers to teach compact students~\cite{fitnet}. Passalis \etal enhance the knowledge distillation process by introducing a concept called feature space probability distribution loss~\cite{passalis2018learning}. Yim \etal propose fast minimization techniques based on 
intermediate feature maps that can also support transfer learning~\cite{yim2017gift}. 
Yin \etal propose DeepInversion that inverts a CNN to reconstruct the dataset for data-free knowledge transfer~\cite{yin2019dreaming}.

\subsection{Dynamic Approaches}
Although static approaches can yield compact architectures with significantly reduced computational cost, these approaches have two distinct disadvantages. First, static approaches permanently remove layers and neurons, thereby decreasing inference capability. Removing rarely activated neurons may be an effective strategy for the average case, but these same neurons may be critical for accurately classifying challenging input instances. Second, statically pruned models process all input instances equally, which may be computationally-inefficient. For example, images from unrelated classes, like 
trucks and birds, have minimal feature overlap: birds do not have headlights and trucks do not have 
wings. As a result, convolutional filters associated with truck features are likely unimportant 
for predicting birds and {\em vice versa}. Furthermore, not all images are equally difficult to 
classify. While difficult images may need deep network embeddings for accurate classification, many 
images may only need shallow embeddings. Hence, the computations associated with deep embeddings are wasted on easy images. 
\\\\
\textbf{Early prediction:}
To address the limitations mentioned above, recent work proposes the concept of conditional computation, where parts of a model are selectively executed depending on the input instance. Early prediction is one type of conditional computation~\cite{adaptivenn,dgate,branchy}. With this approach, predictive signals determine if a set of shallow embeddings is sufficient to correctly classify an input instance. If so, the instance is classified without receiving further processing from deeper layers. Although this approach is an intuitive solution for input-driven efficient inference, early prediction discards all embeddings subsequent to the prediction point. 
\\\\
\textbf{Module selection:}
Other approaches address the above limitation and offer more selection flexibility. For example, BlockDrop, a reinforcement learning-based approach, can learn which arbitrary sets of residual blocks 
to drop in a ResNet architecture~\cite{blockdrop}. Leroux \etal design a ResNet-based network with parameter sharing and an adaptive computation time mechanism to reduce parameters and adaptively execute layers~\cite{iamnn}. Odena \etal propose adaptively constructing computation graphs from sub-modules using a reinforcement learning-based controller~\cite{odena}. Liu \etal propose a new 
type of DNN augmented with control modules to selectively execute subsets of the model with 
Q-learning~\cite{dynamicdnn}. 
\\\\
Despite their additional flexibility over early prediction methods, these approaches have limited dynamics that are restricted to modules. In addition, they typically use computationally-intensive
reinforcement learning to derive a control policy, due to the non-differentiable nature of 
the on/off decisions for each module. A more suitable approach for achieving dynamic selection may increase the flexibility and granularity of dynamic paths rather than be limited to pre-defined modules. Also, training the decision controller jointly with the backbone model in an end-to-end fashion may be favorable. 
\\\\
\textbf{Channel pruning:}
Recent work in channel selection allows greater flexibility for conditional computation. For example, Gao \etal achieve channel-wise selective computation with a feature boosting and suppression mechanism to predictively evaluate feature map salience and skip unimportant channels at run time~\cite{FBS}. However, this method halts the network during the salient channel selection process and introduces overhead. Also, it utilizes a $k$-winners-take-all function to preserve the top $k$ salient channels. It may be difficult to determine the optimal $k$ for each convolutional layer {\em a priori}. 

\section{Methodology}
Feature and layer importance in CNNs varies depending on the input instance. This input-dependence can be exploited in designing efficient networks for energy- and computation-constrained scenarios,
since irrelevant feature maps can, in principle, be ignored without sacrificing accuracy. 
\\\\
We propose a methodology with hierarchical dynamics to achieve on-the-fly selective execution of CNNs for efficient inference. At a coarse-grain level, only salient layers for image discrimination are retained at inference time, while other layers are skipped. At a finer-grain level, only salient 
feature maps/channels associated with retained layers are preserved. This multi-level approach imposes layer-wise and channel-wise sparsity, which significantly reduces computational cost while preserving high classification accuracy.  
\\\\
We primarily focus our attention on constructing a dynamic building block that can adapt its future computation graph based on its input. The block-based approach makes the method readily applicable across various network topologies, given the widespread usage of blocks in networks such as ResNets~\cite{he2016deep}, Inceptions~\cite{inception}, MobileNets~\cite{mobilenet}, ShuffleNets~\cite{shufflenet}, and EfficientNets~\cite{efficientnet}.
\\\\
We hypothesize that an input tensor that contains meaningful information from previous layers can be used to learn execution rules for the current block. In particular, these rules can decide if a block or the channels therein are important for correctly classifying the input instance. To test this hypothesis, we design a block that can selectively execute certain convolutional layers and convolutional channels within these layers, depending on the input instance. 

\subsection{L-Net: Dynamic Layer Skipping for Depth Flexibility}
\begin{figure}[t]
\begin{center}
\includegraphics[width=5.5cm]{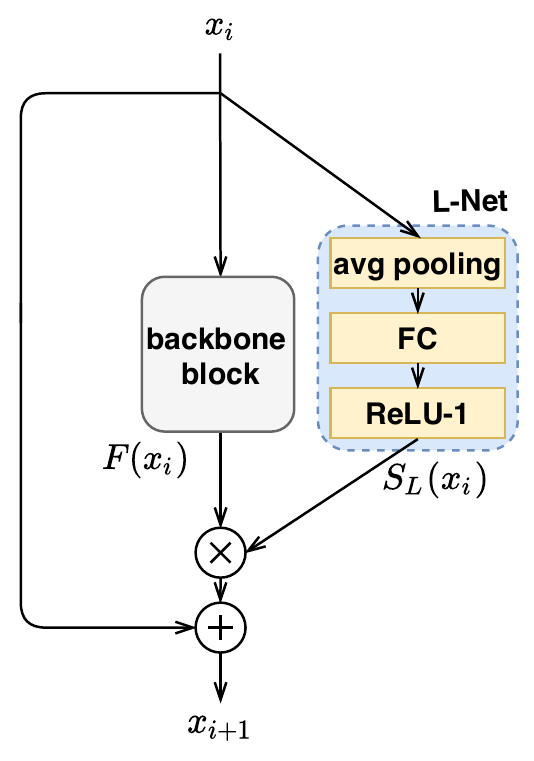}
\end{center}
\caption{Schematic diagram of L-Net.}
\label{l-net}
\vspace{-3mm}
\end{figure}

Not all input instances require all layer-wise computations to be correctly classified~\cite{dgate}.
In modern DNNs, repeated blocks are built on top of each other to fine-tune feature details. Harder 
samples may need deeper embeddings to be accurately classified, while easier samples may only need 
shallow embeddings. In other words, shallower inference is viable for easier samples, while deeper layers are needed for harder cases to maintain performance. 
\\\\
We propose a depth-wise skipping framework that dynamically selects salient layers needed for high 
classification performance. Most modern DNNs have adopted a block-based residual learning design following the remarkable success of ResNet~\cite{he2016deep} (for example, MobileNetV2~\cite{mobilenetv2}, ShuffleNet~\cite{shufflenet}, and ResNext~\cite{resnext}), which solved the accuracy degradation problem associated with DNNs. We therefore construct our skipping methodology to be generally applicable to arbitrary blocks used in modern DNNs.
\\\\
To enable dynamic block skipping, we propose adding a small network called L-Net to arbitrary blocks with shortcut connections. The L-Net architecture is illustrated in Fig.~\ref{l-net}. L-Net contains three parts: a global average pooling layer, a fully-connected layer, and a ReLU-1 activation function. We designed L-Net to be small and shallow to make sure that the additionally introduced computations and parameters are negligible compared to the original backbone neural network. 
\\\\
For simplicity, consider an arbitrary $i^{th}$ building block with skip connections in a general block-based backbone DNN. The input to the $i^{th}$ block is denoted as $x_i$, the output is $x_{i+1}$, and the function represented by the block is $F: \mathbb{R}^{C_{i} \times H_{i} \times W_{i}} \rightarrow \mathbb{R}^{C_{i+1} \times H_{i+1} \times W_{i+1}}$. The transformation performed by the original block architecture can be represented as follows: 
\begin{equation}
    x_{i+1}= F(x_i) + x_i
\end{equation}
where $F$ contains one or more convolutional layers, which account for the majority of network
computations. In order to dynamically skip unnecessary blocks and thereby reduce computational cost, 
an L-Net is added in parallel to each block. For the $i^{th}$ block, L-Net takes $x_i$ as input. 
Within L-Net, $x_i$ first passes through a global average pooling layer to reduce the spatial size 
to 1 per channel. Next, the output vector is passed to a fully-connected layer $FC$ followed by a 
ReLU-1 activation, which outputs a block salience score, denoted by $S_L$, between 0 and 1. The 
ReLU-1 output is a scaling factor applied to the block output, $F(x_i)$. If the block salience is 
zero, the block is skipped. We formalize the L-Net controlled block as follows:

\begin{equation}
    x_{i+1}= F(x_i) \cdot S_L(x_i) + x_i
\end{equation}

\begin{equation}
    S_L(x_i) = \mbox{ReLU-1}(\:FC\:(\:\mbox{global-avg-pool}\:(x_i)\:)\:) 
\end{equation}

\textbf{}
\subsection{C-Net: Dynamic Channel Selection for Width Flexibility}
\begin{figure}[t]
\begin{center}
\includegraphics[width=7.9cm]{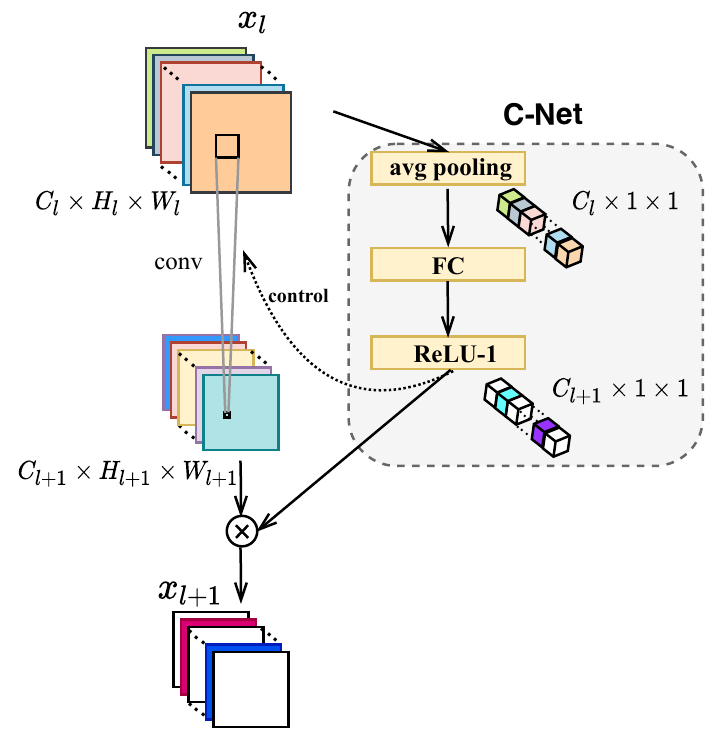}
\end{center}
\caption{Schematic diagram of C-Net.}
\label{c-net}
\vspace{-3mm}
\end{figure}

Most of the DNN computational cost is incurred in the convolutional layers. Within these layers, the contribution of individual channels is highly input-dependent \cite{FBS}. For example, feature maps for cars are not useful for classifying horses, since these feature maps are generally not activated post-ReLU. As a result, these irrelevant feature maps may be avoided without degrading classification accuracy. 
\\\\
While L-Net improves computational efficiency through block-level skipping, the overall 
computational cost can be further reduced by exploiting image-dependent salience differences at the 
channel level within retained blocks. To this end, we propose a complementary approach to L-Net, 
called C-Net, which dynamically prunes unimportant channels in an input-driven manner. 
\\\\
Generally, consider the $l^{th}$ convolutional layer $g_l$ of a deep CNN. Denote the mapping performed 
by the $l^{th}$ layer as $g_l: \mathbb{R}^{C_{l} \times H_{l} \times W_{l}} \rightarrow 
\mathbb{R}^{C_{l+1} \times H_{l+1} \times W_{l+1}}$, which computes feature maps $x_{l+1} \in 
\mathbb{R}^{C_{l+1} \times H_{l+1} \times W_{l+1}} $ given input $x_l \in \mathbb{R}^{C_{l} \times 
H_{l} \times W_{l}}$.  The goal of C-Net is to predict channel salience and only execute a subset of corresponding important convolutions within the total $c_{l+1}$ convolutions in the layer. 
\\\\
The schematic of C-Net is shown in Fig.~\ref{c-net}. C-Net is added in parallel to the $l^{th}$ 
layer. Like L-Net, C-Net is a compact network that contains a global average pooling layer, a 
fully-connected layer, and a ReLU-1 activation function. The fully-connected layer is designed to have $C_{l+1}$ units to match the output's number of channels. C-Net shares the same input $x_l$ as the $l^{th}$ layer. Within C-Net, the global average pooling layer processes the input and produces a vector of length $C_l$. Next, this vector goes through the fully-connected layer and a ReLU-1 activation to produce a channel salience score of size $C_{l+1}$, denoted by $S_C(x_l)$, between 0 and 1. The dynamic layer with channel selection can be expressed as follows:

\begin{equation}
    g_l \cdot S_C(x_l) = g_l \cdot  \mbox{ReLU-1}(\:FC\:(\:\mbox{global-avg-pool}\:(x_l)\:)\:) 
\end{equation}

\par \medskip \noindent
where $S_C(x_l)^k \in \mathbb{R} $, $k \in \{1,...,C_{l+1} \}$ is the salience score for the 
$k^{th}$ channel, which is multiplied with all elements of the $k^{th}$ channel. During inference, convolutions are not executed if their associated channels have a salience score of 0. Channels with a non-zero score are calculated, and their resulting feature maps are scaled by their corresponding score $S_C(x_l)$.

\subsection{Joint Design: LC-Net}
\begin{figure*}[h]
\centerline{
\includegraphics[scale=1.25]{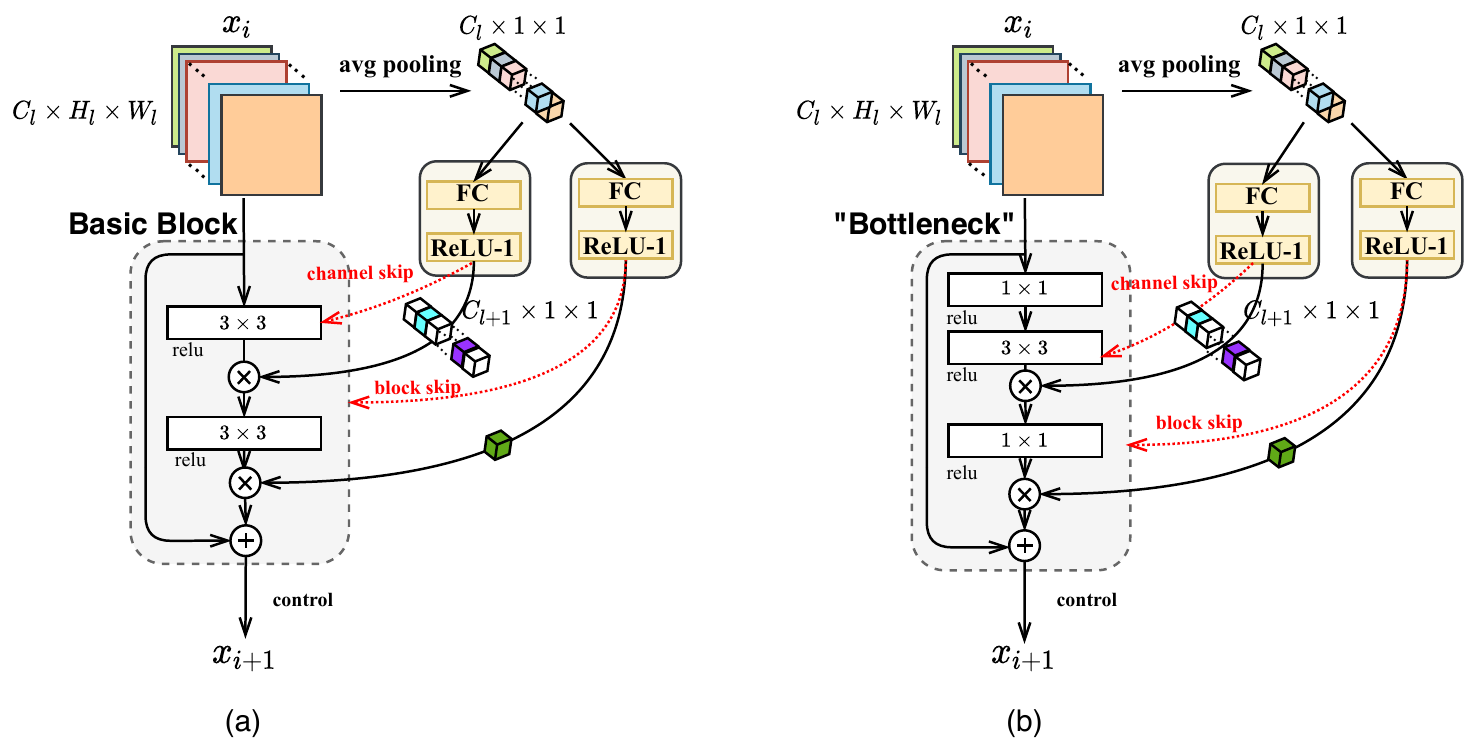}}
\caption{Illustration of joint LC-Net design: (a) basic block augmented with LC-Net, and 
(b) ``bottleneck" block augmented with LC-Net.}
\label{jointdesign}
\end{figure*}

L-Net and C-Net are orthogonal approaches that, respectively, enable depth-wise and channel-wise 
skipping and scaling for efficient dynamic inference. As such, we integrate the two approaches to 
achieve fully dynamic inference and minimize computational cost. 
\\\\
For concreteness, we illustrate an instantiation of our co-design methodology on the ResNet family. Within this family, there are two types of building blocks: basic and $``$bottleneck.$"$ A basic block consists of two 3$\times$3 convolutional layers, and a $``$bottleneck$"$ block consists of a sequence of 1$\times$1, 3$\times$3, and 1$\times$1 convolutional layers. The joint design structures for both 
types of building blocks are illustrated in Fig.~\ref{jointdesign}. L-Net and C-Net are both added in parallel to either building block type and take the same input. Since L-Net and C-Net both use a global average pooling process, we are able to reduce computation overhead by sharing one global average pooling layer between the two. As mentioned earlier, the block salience score produced by L-Net is multiplied with the residual mapping. The channel salience score is multiplied with the output of the first layer within the building block. For the basic block, this corresponds to the first 3$\times$3 convolutional layer; for the $``$bottleneck$"$ block, this corresponds to the first 1$\times$1 convolutional layer. 
\\\\
In addition to sharing a pooling layer, we introduce two design choices to encourage training convergence and allow on-the-fly dynamics with minimal latency overhead.

\begin{figure}[h]
\centering
\includegraphics[width=\columnwidth]{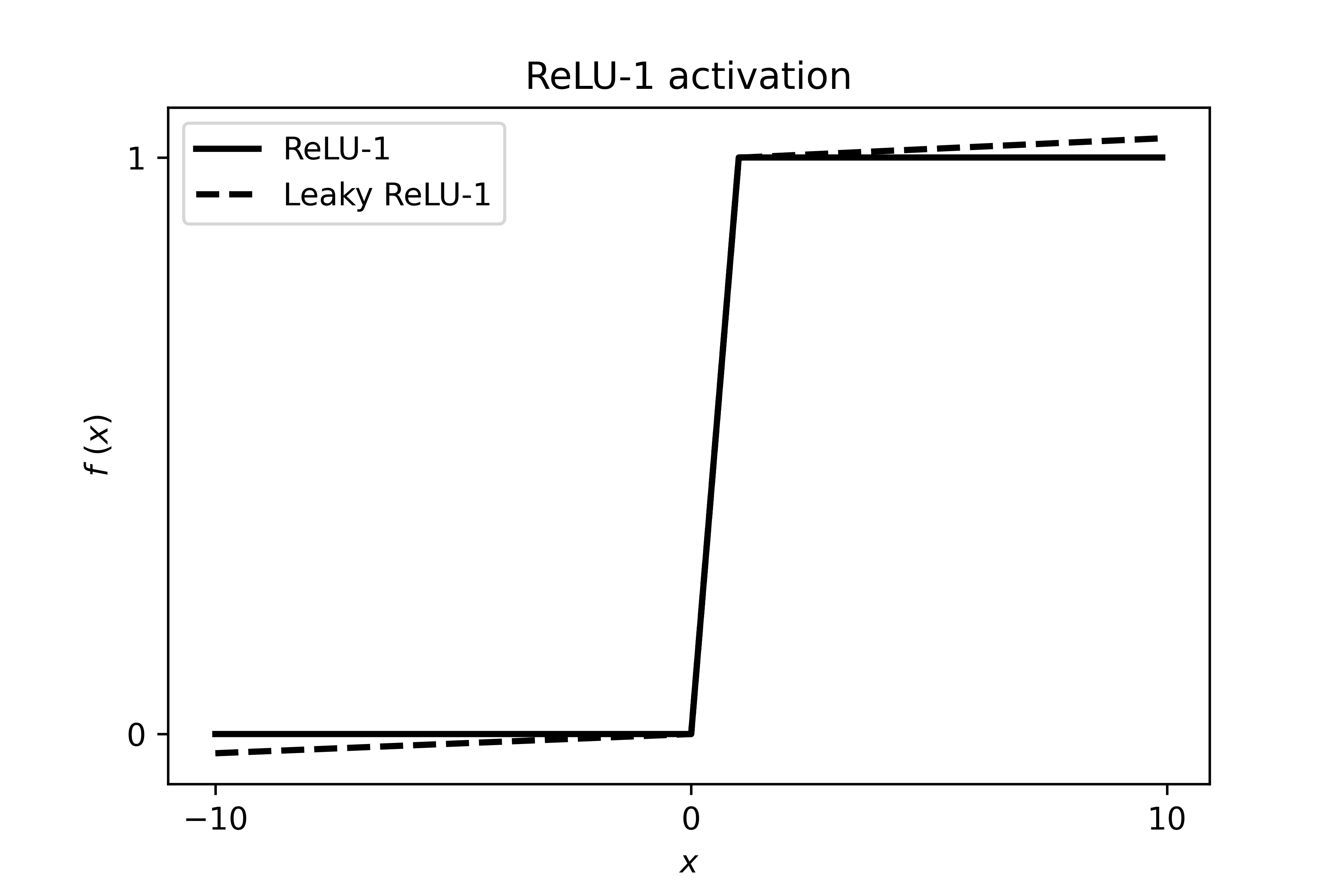}
\caption{Visualization of the ReLU-1 activation. A leaky ReLU-1 is used for training while a standard 
ReLU-1 is used during inference.}
\label{plotrelu1}
\end{figure}

\begin{subsubsection}{ReLU-1}
In both L-Net and C-Net, we obtain a salience score between 0 and 1 via a ReLU-1 activation function, 
which is displayed in Fig.~\ref{plotrelu1} and formulated as:

\begin{equation}
  \mbox{ReLU-1}(x) = 
  \begin{cases} 
      0 & x\leq 0 \\
      x & 0 < x\leq 1 \\
      1 & x > 1  
  \end{cases}  
\end{equation}
\\\\
During training, we use a leaky ReLU-1 to encourage convergence. Unlike the sigmoid activation 
function, which also produces results between 0 and 1, this ReLU-1 activation function does not suffer from vanishing gradients during training and is able to produce a strict range of values between 0 and 1 at inference time. In addition, the leaky ReLU-1 is less prone to exploding activations for positive inputs than the standard ReLU. Since the ReLU-1 function is differentiable, LC-Net and the backbone model can be jointly trained in an end-to-end fashion, unlike reinforcement learning-based policy controllers. This improves training efficiency.

\end{subsubsection}
\begin{subsubsection}{Parallelism}
We designed L-Net and C-Net to be parallel to the building block such that the control nets and the backbone network can all execute 
simultaneously. L-Net and C-Net have fewer computations than the first convolutional layer in the 
main building block. Hence, these two networks can produce salience scores before the first  convolutional layer has finished execution. This leads to memory and cache fetching efficiency for 
the next convolutional layer via filter-level suppression. Therefore, unless L-Net predicts that the whole block can be skipped, some of the next convolutional layer's computations can be saved based on which channel saliencies are set to zero. 
\end{subsubsection}

\section{Experimental Results}
We run extensive experiments on classification tasks with CIFAR-10~\cite{cifar} and 
ImageNet~\cite{imagenet} datasets to demonstrate the effectiveness of our fully dynamic inference 
framework. We instantiate the fully dynamic design on the ResNet family backbone in PyTorch~\cite{pytorch}. On both datasets, we compare FLOPs and accuracy of our method with related work in dynamic inference. In addition, we visualize the per-instance dynamics during inference to better understand our framework's behavior.

\subsection{Datasets}
We evaluate our method on two popular datasets: CIFAR-10 and ImageNet. Both datasets consist of
colored natural images. CIFAR-10 is divided into 50,000 training instances and 10,000 testing 
instances of resolution 32$\times$32 labeled for 10 classes. To demonstrate the scalability of our method to larger and more complex datasets, we evaluate it on ImageNet, which consists of images of size 224$\times$224 labeled for 1000 classes. ImageNet consists of 1.2M training images and 50,000 validation images.

\subsection{Training and Implementation Details}
We adopt stochastic gradient descent (SGD) with a Nesterov momentum of 0.9 without damping to train all models. We also use a weight decay (L2 penalty) of 0.0005. All backbone models with L-Net and C-Net added are trained from scratch. On CIFAR-10, we train the model for 270 epochs. We set the initial learning rate to 0.01 and decay it by ten-fold every 90 epochs. The training batch size we use on CIFAR-10 is 96. For training on ImageNet, we adjust the batch size to 256. We also train the backbone model and L-Net/C-Net with separate learning rates. Specifically, we set the initial learning rate to 0.005 for the backbone model and 0.00001 for the added L-Net and C-Net. Both learning rates decay ten-fold every 30 epochs for a total number of 120 epochs. After training the proposed dynamic framework, we run inference and record the classification accuracy and FLOPs. 

\subsection{Main Results}
\begin{figure}[h]
\centering
\includegraphics[width=\columnwidth]{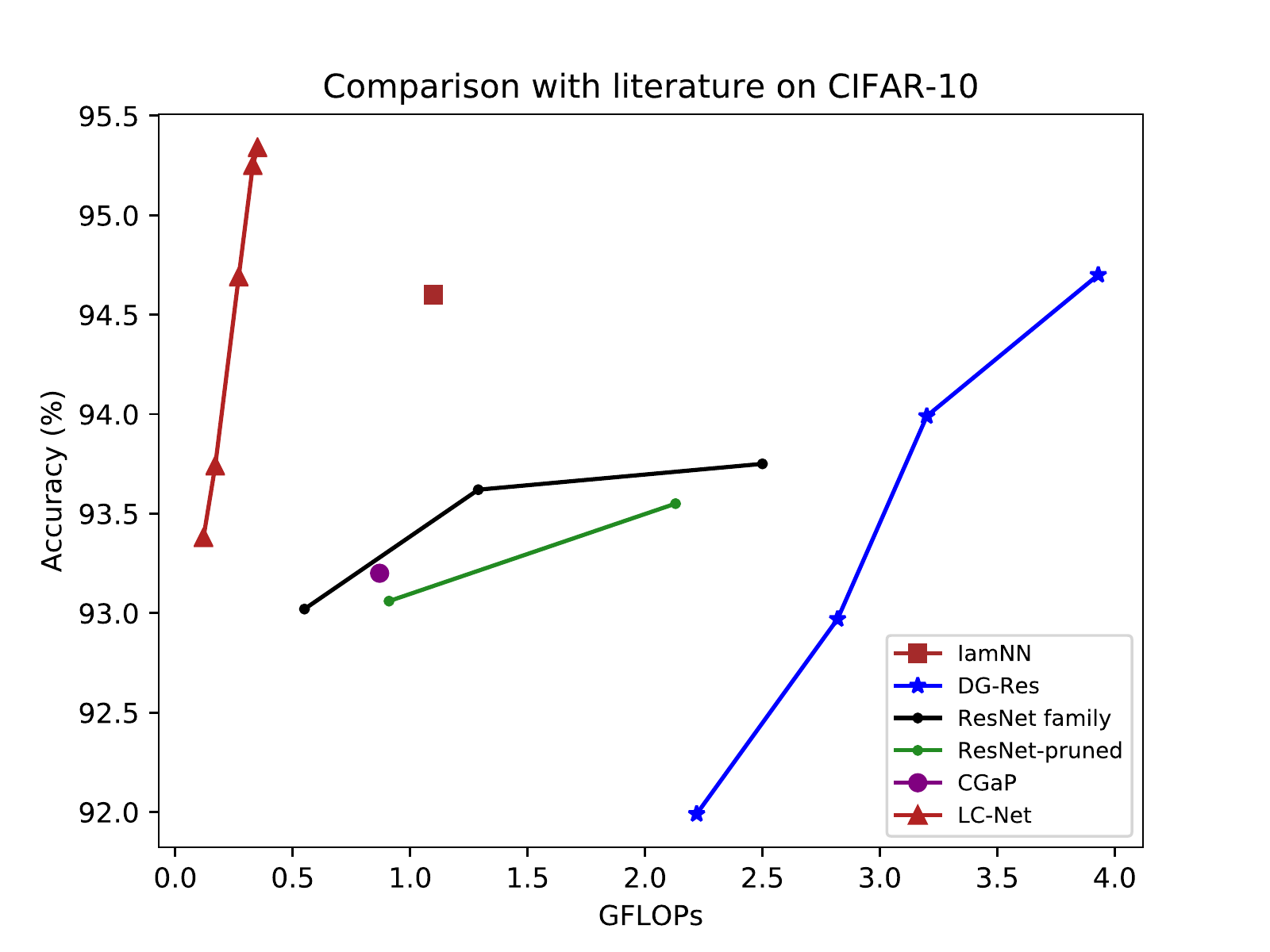}
\caption{Comparison of our proposed method with the literature on the CIFAR-10 dataset.
Top-left is better. Our framework outperforms all previous dynamic methods.}
\label{plotcifar}
\end{figure}

We compare our methodology to existing methods for efficient inference. The methods we consider include IamNN~\cite{iamnn}, Decision Gate-Resnet (DG-Res)~\cite{dgate}, Continuous Growth and Pruning 
(CGaP)~\cite{cgap}, filter pruning~\cite{prunefilter}, and Neuron Importance Score Propagation (NISP)~\cite{nisp}. We use FLOPs and percent accuracy to assess model performance.
\\\\
\textbf{CIFAR-10:} As shown in Table~\ref{tb:cifar_results}, our method outperforms all prior art 
with respect to accuracy and FLOPs on CIFAR-10. Compared to IamNN, our method, which uses a ResNet-18
backbone, achieves $0.65\%$ higher accuracy with more than $3\times$ fewer FLOPs. Compared to 
DG-Res, our model has $3.26\%$ higher accuracy and more than $6.7\times$ fewer FLOPs than the most compact configuration, and $0.55\%$ higher accuracy and more than $11.9\times$ fewer FLOPs than the highest accuracy configuration. Compared to CGaP, which has the fewest FLOPs among the methods we considered, we achieve $2.05\%$ higher accuracy and $0.54$ fewer GFLOPs. Similar to the trend shown for DG-Res, our method is more accurate and has fewer FLOPs than filter-pruned ResNet-56 and ResNet-110 models. These results are depicted graphically in Fig.~\ref{plotcifar}.

\begin{table}[h!]
\begin{center}
\caption{Comparison of our proposed method with the literature on the CIFAR-10 dataset.} 
\label{tb:cifar_results}
\begin{tabular}{ |c|c|c|c| } 
\hline
Dataset & Network & GFLOPs & Acc(\%) \\
\hline
\multirow{12}{4.5em}{CIFAR-10} & ResNet101~\cite{he2016deep} &	2.50 &	93.75 \\
& Res50 ~\cite{he2016deep}	& 1.29	& 93.62 \\ 
& Res18~\cite{he2016deep} & 0.55	& 93.02 \\
& IamNN~\cite{iamnn}	& 1.10 & 94.60 \\
& DG-Res~\cite{dgate} - config A & 2.22	& 91.99 \\
& DG-Res~\cite{dgate} - config B & 2.82	& 92.97 \\
& DG-Res~\cite{dgate} - config C	& 3.20	& 93.99 \\
& DG-Res~\cite{dgate} - config D	& 3.93	& 94.70 \\
& CGap~\cite{cgap} & $>0.87$ & 93.20 \\
& ResNet-110-pruned~\cite{prunefilter} & 2.13 & 93.55 \\
& ResNet-56-pruned~\cite{prunefilter} & 0.91 & 93.06 \\
& \textbf{LC-Net} & \textbf{0.33} & \textbf{95.25} \\
& \textbf{LC-Net pre-trained (parallel)} & \textbf{0.19} & \textbf{93.27}\\ 
& \textbf{LC-Net pre-trained (sequential)} & \textbf{0.06} & \textbf{93.27}\\ 
\hline
\end{tabular}
\end{center}
\end{table}

\par \medskip \noindent
In addition to training LC-Net with our backbone model from scratch, we also experimented with
augmenting a pre-trained backbone with LC-Net. In principle, using a pre-trained backbone can save 
considerable training time. To encourage sparser dynamic executions, we add L1 regularization to both prediction tensors coming out of L-Net and C-Net for each building block, namely, $S_L$ and $S_C$. 
\\\\
We show results for a ResNet-18 backbone augmented with LC-Net. As can be seen from Table~\ref{tb:cifar_results}, using a  pre-trained model with L1 regularization results in an
additional dramatic reduction in FLOPs with only a slight reduction in accuracy compared to the
LC-Net model trained from scratch. Using a pre-trained backbone, we also tested two different
configurations that allowed us to explore the trade-off between FLOPs and inference
latency. In the on-the-fly predictive configuration we discussed earlier, LC-Net produces
predictive salience scores in parallel with the computations within the main block, which
incurs no latency overhead. In the second configuration, LC-Net is placed in series with
the main block such that predictive salience scores are generated before the main block
executes. Our results suggest that for very sparse models, the series configuration yields a
substantial reduction in FLOPs, despite a possible increase in latency, compared to
our original parallel configuration. Therefore, our framework offers the flexibility to choose between different configurations that can minimize latency or minimize computational cost depending on the deployment constraints. 
\\\\
\textbf{ImageNet:} For the ImageNet dataset, we compare our results on ResNet-50 to its baseline, as well as IamNN and NISP. Compared to the baseline, our method has comparable Top-1 and Top-5 accuracy 
($74.1\%$ and $92.1\%$, respectively) while significantly reducing GFLOPs to $2.89$ ($1.42\times$ 
reduction). Consistent with our finding for a ResNet-18 backbone on CIFAR-10, our method has
greater accuracy and fewer FLOPs compared to IamNN and NISP-50-A, as shown in Table
\ref{tb:imagenet_results}.  These results demonstrate the scalability of our method to larger, more 
complex datasets like ImageNet. 

\begin{table}[h!]
\begin{center}
\caption{Comparison of our proposed method with the literature on the ImageNet dataset.} 
\label{tb:imagenet_results}
\begin{tabular}{ |c|c|c|c|c| } 
\hline
Dataset & Models & GFLOPs & Top-1 Acc.(\%) & Top-5 Acc.(\%) \\
\hline
\multirow{4}{4.5em}{ImageNet} 
& ResNet-50 [1] & 4.09  &	75.3 & 92.2   \\
& IamNN [2]    & 4.00	&   69.5 & 89.0   \\ 
& NISP-50-A [6] & 2.97	&   72.8 & -      \\
& \textbf{LC-Net} & \textbf{2.89} & \textbf{74.1} & \textbf{92.1}\\
\hline
\end{tabular}
\end{center}
\end{table}

\subsection{Qualitative Analysis}
Next, we assess the qualitative behavior of our dynamic framework to validate our intuitions.

\subsubsection{Relationship between instance complexity and FLOPs}
\begin{figure*}[t]
\centerline{
\includegraphics[scale=0.33]{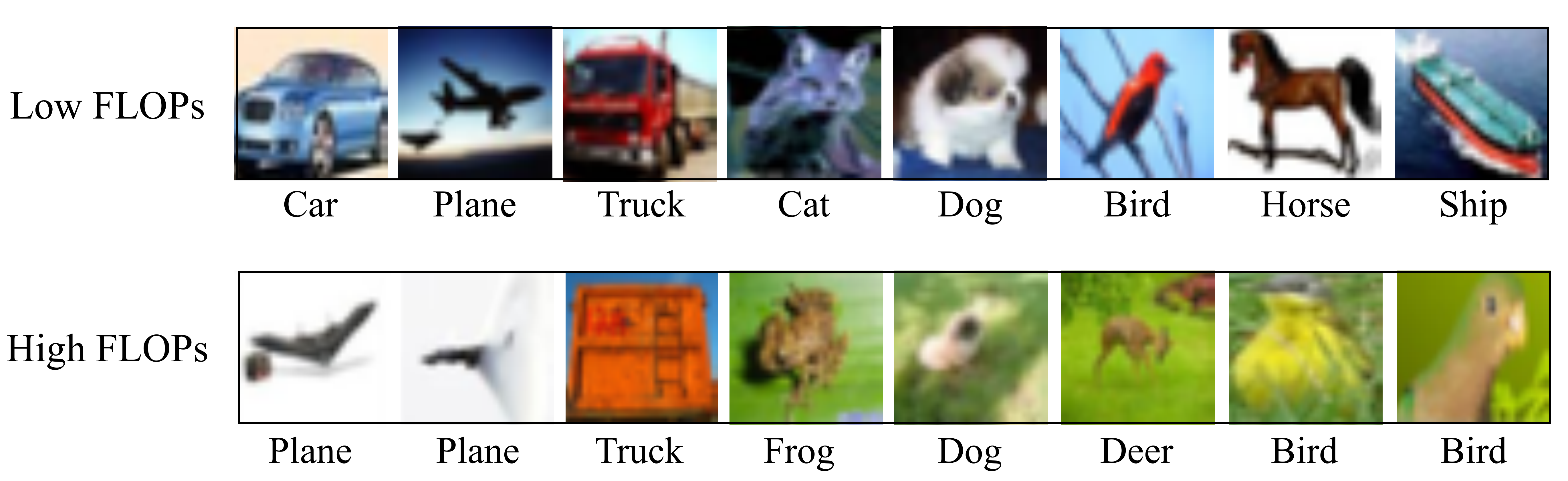}}
\caption{Examples of images that require low (first row) and high (second row) FLOPs. The label under 
each image is the ground truth class.}
\label{flopsplot}
\end{figure*}

Our study was initially motivated by the intuition that images of different complexities require 
different amounts of computation. In particular, easily classifiable images should require fewer 
deep embeddings than more complex or atypical images. In addition, we hypothesized that since 
convolutional filters capture class-specific information, not all convolutional channels should be computed for each input instance. 
\\\\
To qualitatively evaluate these intuitions and gain a better understanding of how our dynamic 
framework behaves during inference, we record FLOPs for each test instance and select representative examples with low and high FLOPs. These examples are shown in Fig.~\ref{flopsplot}. Images with low and high FLOPs are shown in the top and bottom rows, respectively. The ground truth labels are shown below each image. 
\\\\
Visual differences are apparent between these two groups. In general, objects in the 
low-FLOPs image group are easily discernible, but images in the high-FLOPs image group are 
more challenging to classify. The low-FLOPs group tends to contain images that are more representative of their corresponding classes. For example, animals have typical poses, while vehicle features like wheels and windshields are clearly visible. Images in the high-FLOPs group, on the other hand, have ambiguous contours and atypical features. For example, the left two images of a plane are not obviously distinct from birds. The truck image shows the back of a truck, while most images in the truck class show the front or side of trucks. For the remaining instances in the high-FLOPs group, the animals have blurry outlines and have low contrast with respect to their background, which increases the likelihood of mistaken identification. 
\\\\
Altogether, these results are 
consistent with our hypothesis that images of different complexity require different amounts of computation. In particular, it appears that easily discriminated or highly class-representative 
images tend to require fewer computations, while more atypical or ambiguous images tend to require more computations. 

\subsubsection{Dynamic Selection}

\begin{figure*}[t]
\centerline{
\includegraphics[scale=0.35]{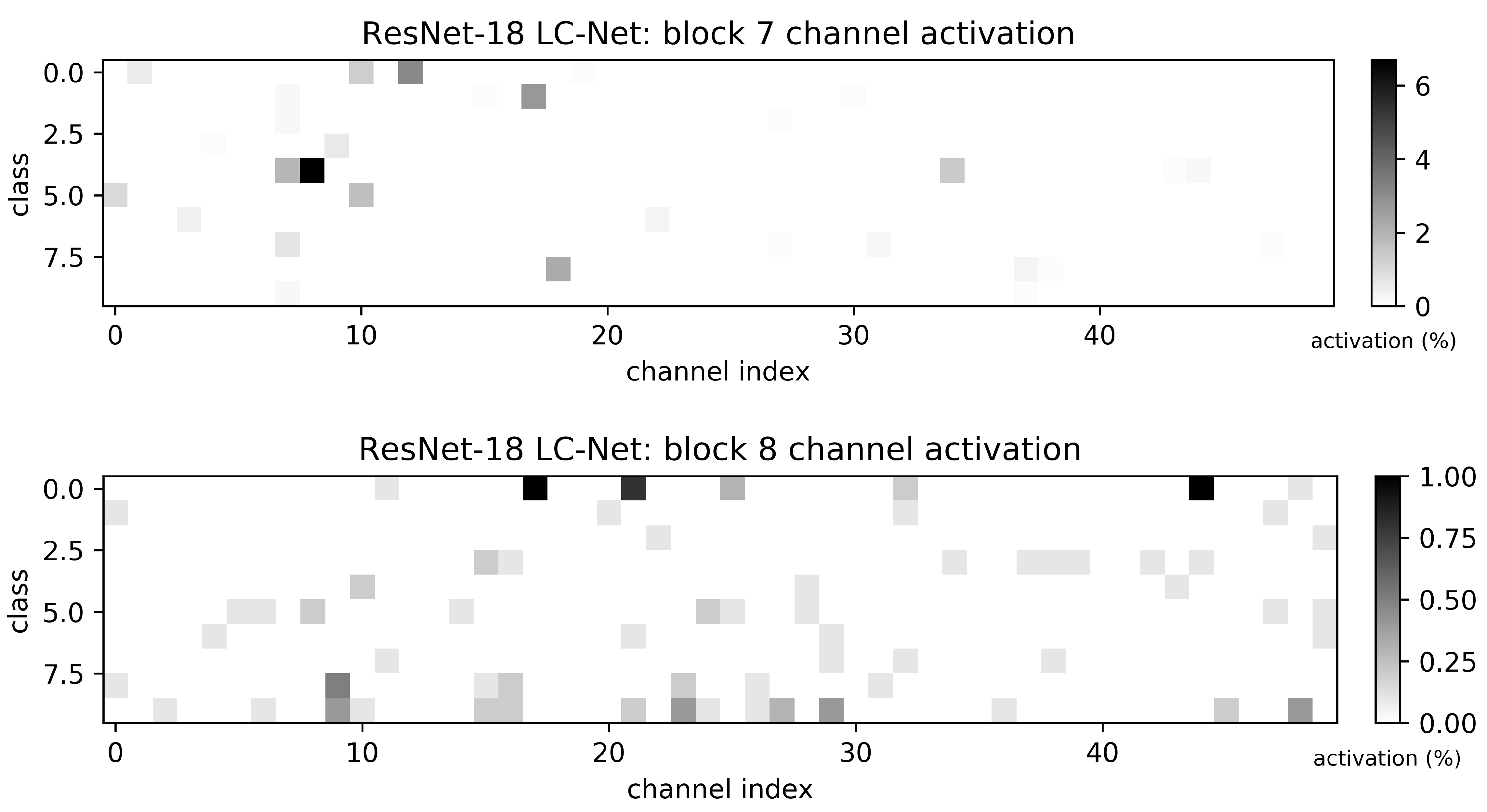}}
\caption{Visualization of channel selection on a ResNet-18 backbone. Each column shows the percentage 
of times a particular channel is executed across image classes.}
\label{dynamicplot}
\end{figure*} 

Next, we visualize the dynamic execution behavior of our approach during inference. We show examples
of dynamic channel execution behavior within the last two basic blocks (namely, block 7 and block 8) of a ResNet-18 based model on CIFAR-10. In Fig.~\ref{dynamicplot}, the color patches indicate the 
activation percentage of each channel for each image class. More specifically, the color patch at 
location ($i,j$) in the figure represents the percentage of image instances in class $i$ that 
activates channel $j$. For visual clarity, only the first 50 channels in each block's second 
3$\times$3 convolutional layer are displayed. As can be seen from the figure, different classes 
require different subsets of channels for accurate classification. In addition, the prevalence of white space, which denotes class-specific channel inactivation, illustrates the sparsity of the dynamic inference model. Furthermore, for the channels that are active for certain classes, the activation percentage is low, implying that most images within these classes do not require the channels' computations. These observations are consistent with the substantially reduced FLOPs 
reported in Table \ref{tb:cifar_results}. 

\section{Conclusions}
We presented a new end-to-end training framework that achieves instance-based, fully dynamic
inference to automatically optimize computational paths within a DNN. Two shallow networks,
L-Net and C-Net, respectively, contribute on-the-fly layer-wise and channel-wise skipping and scaling 
decisions. Our experiments with CIFAR-10 and ImageNet demonstrate that our selective execution 
approach results in a dramatic reduction in FLOPs and substantially higher accuracy than competing 
dynamic inference methods. 

\ifCLASSOPTIONcaptionsoff
  \newpage
\fi

\bibliographystyle{IEEEtran} 
\bibliography{arxiv}

\end{document}